# Machine Learning Models for Reinforced Concrete Pipes Condition Prediction: The State-of-the-Art Using Artificial Neural Networks and Multiple Linear Regression in a Wisconsin Case Study

Mohsen Mohammadagha[1,*], Mohammad Najafi[1], Vinayak Kaushal[1,*], Ahmad Mahmoud Ahmad Jibreen[2]

[1] Center for Underground Infrastructure Research and Education (CUIRE), The University of Texas at Arlington, Arlington, TX 76013, USA; najafi@uta.edu
[2] Assistant professor at the university of Texas at Tyler, ahmadmahmoudahm.jibreen@mavs.uta.edu

[*] **Correspondence**: mxm4340@mavs.uta.edu (M.M); vinayak.kaushal@uta.edu (V.K)

**Abstract:**

The aging sewer infrastructure in the U.S., covering 2.1 million kilometers, encounters increasing structural issues, resulting in around 75,000 yearly sanitary sewer overflows that present serious economic, environmental, and public health hazards. Conventional inspection techniques and deterministic models do not account for the unpredictable nature of sewer decline, whereas probabilistic methods depend on extensive historical data, which is frequently lacking or incomplete. This research intends to enhance predictive accuracy for the condition of sewer pipelines through machine learning models—artificial neural networks (ANNs) and multiple linear regression (MLR)—by integrating factors such as pipe age, material, diameter, environmental influences, and PACP ratings. ANNs utilized ReLU activation functions and Adam optimization, whereas MLR applied regularization to address multicollinearity, with both models assessed through metrics like RMSE, MAE, and $R^2$. The findings indicated that ANNs surpassed MLR, attaining an $R^2$ of 0.9066 compared to MLR's 0.8474, successfully modeling nonlinear relationships while preserving generalization. MLR, on the other hand, offered enhanced interpretability by pinpointing significant predictors such as residual buildup. As a result, pipeline degradation is driven by pipe length, age, and pipe diameter as key predictors, while depth, soil type, and segment show minimal influence in this analysis. Future studies ought to prioritize hybrid models that merge the accuracy of ANNs with the interpretability of MLR, incorporating advanced methods such as SHAP analysis and transfer learning to improve scalability in managing infrastructure and promoting environmental sustainability.

**Keywords**:
Machine Learning, Model Prediction, Sewer Pipe, linear regression, Condition Assessment.

# 1. Introduction:

Urban drainage systems are essential components of modern infrastructure, designed to collect and transport wastewater efficiently to treatment facilities. In the United States alone, approximately 2.1 million kilometers of sewer lines exist, many of which are aging and deteriorating due to environmental and operational stresses (Sterling et al., 2010). As these systems age, their structural and hydraulic capacities diminish, leading to potential failures that can result in urban flooding, groundwater contamination, and public health hazards (Djordjević et al., 2005; Rutsch et al., 2008). The Environmental Protection Agency (EPA) estimates that deteriorated sewer systems contribute to up to 75,000 sanitary sewer overflows annually, with significant economic and environmental consequences (EPA, 2024). Consequently, effective asset management strategies are essential to ensure the longevity and functionality of these systems while minimizing environmental and economic impacts.

The deterioration of sewer pipelines is influenced by a complex interplay of physical, operational, and environmental factors. Key contributors include pipe age, material type, diameter, depth, soil corrosivity, groundwater levels, and operational conditions such as flow velocity and sediment accumulation (Ana et al., 2009; Davies et al., 2001). For instance, reinforced concrete pipes are highly susceptible to sulfide-induced corrosion in low-flow conditions, while plastic pipes may experience deformation under heavy loads (Salman & Salem, 2012). These factors not only accelerate structural degradation but also increase the likelihood of hydraulic inefficiencies and blockages.

Traditional methods for assessing sewer pipe conditions often rely on visual inspections or deterministic models. While visual inspections using closed-circuit television (CCTV) remain the most common approach for identifying defects such as cracks or root intrusions, they are labor-intensive, subjective, and prone to inconsistencies (Harvey & McBean, 2014). Deterministic models attempt to predict pipe deterioration based on predefined relationships between variables but fail to account for the stochastic nature of degradation processes (Ana & Bauwens, 2010; Dirksen et al., 2007). Probabilistic models such as Markov chains and survival analysis have been developed to address these limitations by incorporating stochastic elements into condition predictions (Baur & Herz, 2002; Micevski et al., 2002). However, these models often require extensive historical data for calibration—data that is frequently unavailable or incomplete in real-world scenarios (Scheidegger et al., 2011a)

Recent advancements in machine learning (ML) offer transformative potential for sewer condition prediction by leveraging large datasets and uncovering complex patterns among explanatory variables. ML techniques such as artificial neural networks (ANNs), gradient boosting trees (GBT), fuzzy logic systems, and adaptive neuro-fuzzy inference systems (ANFIS) have demonstrated their ability to model nonlinear relationships between deterioration factors and pipeline conditions (Mohammadi et al., 2020). For example, ANN-based models have been employed to estimate the remaining useful life (RUL) of pipelines by analyzing variables such as pipe material, age, diameter, and environmental factors like soil type and groundwater levels (Tavakoli et al., 2020). Similarly, GBT models have shown high accuracy in predicting structural conditions by integrating diverse datasets from inspection records and environmental databases (Mohammadi et al., 2020).

The integration of ML with comparative factor analysis represents a significant advancement in predictive modeling for sewer infrastructure. Comparative factor analysis enables researchers to identify the most influential variables affecting pipeline deterioration while reducing data collection costs by focusing on key predictors. Studies have consistently highlighted pipe age as the most critical factor influencing deterioration rates across various datasets (Hawari et al., 2018; Malek Mohammadi et al., 2020). Other significant variables include pipe material—where concrete pipes are particularly vulnerable to sulfate attack—and operational factors such as flow velocity that exacerbate sediment deposition or erosion processes (Davies et al., 2001).

Probabilistic methods such as Monte Carlo simulations (Abuhishmeh & Hojat Jalali, 2023) further enhance ML-based frameworks by modeling uncertainties associated with deterioration processes. These methods allow for the evaluation of various failure modes at both serviceability and ultimate limit states by incorporating stochastic variations in material properties, environmental factors, and structural lads(Elmasry et al., 2016). For instance, Micevski et al. (2002) developed and validated a Markov model using Bayesian techniques and MCMC for stormwater pipe deterioration, identifying key factors like diameter, material, soil type, and exposure classification (Micevski et al., 2002).

Deep learning approaches have also emerged as powerful tools for automating defect detection in sewer inspections. Convolutional neural networks (CNNs) have been used to classify defects such as cracks or root intrusions from CCTV images with high accuracy and consistency compared to manual inspections (S. S. Kumar et al., 2018). Advanced object detection algorithms like Faster R-CNN and YOLOv3 enable real-time localization of defects within video frames while maintaining computational efficiency (Cheng & Wang, 2018)

Despite these advancements, significant challenges remain in developing robust models that can generalize across diverse sewer networks with varying material compositions, installation practices, and environmental exposures. For example, Horold and Baur noted that selective survival biases in historical datasets can lead to overestimation of pipe lifespans when early failures are excluded from the analysis. Their work, along with other researchers, contributed to the development of tools such as AQUA-WertMin and DynaStrat for modeling sewer network deterioration (Hörold & Baur, 2000). Furthermore, discrepancies in inspection intervals can introduce systematic errors into model calibrations. Addressing these challenges requires integrating diverse datasets from multiple sources, including soil databases, weather records, and inspection logs—into unified predictive frameworks.

This study aims to address these challenges by integrating comparative factor analysis with machine learning models for predicting sewer pipe conditions. The proposed methodology combines probabilistic modeling with advanced ML algorithms to analyze multiple failure modes under varying environmental conditions. Case studies from Wisconsin demonstrate the practical application of this framework in real-world scenarios characterized by diverse soil types and climatic conditions. By incorporating explanatory variables such as chloride concentration gradients or reinforcement ductility losses into predictive models, this research seeks to develop a comprehensive tool for sewer asset management.

In summary, this research represents a significant step forward in leveraging machine learning technologies for infrastructure management. By integrating comparative factor analysis with advanced predictive algorithms like ANNs or GBTs into condition assessment frameworks, it aims to provide utility managers with actionable insights into sewer pipe deterioration processes. These innovations hold the potential to guide inspection planning more effectively while optimizing rehabilitation efforts—ultimately extending the service life of critical infrastructure assets while ensuring public safety and environmental sustainability.

## 2. Literature review:

Sewer systems are among the basic units of urban infrastructure designed to convey wastewater to treatment. Aging infrastructure, environmental stresses, and operational challenges are the reasons for progressive deteriorations. In this regard, effective condition assessment models are necessary for maintenance and rehabilitation prioritization to prevent economic losses, environmental damages, and public health crises. (ASCE, 2017; Malek Mohammadi et al., 2020). The increasing complexity of urban systems necessitates advanced deterioration models that integrate physical, environmental, and operational factors.

### 2.1 Factors Influencing Sewer Deterioration

#### 2.1.1 Physical Factors

Sewer deterioration is influenced by several critical factors, including age, material, diameter, length, and depth. Age is a key determinant, as older pipes experience higher defect rates due to material fatigue, length, and prolonged exposure to environmental stressors. This phenomenon is often illustrated by the "bathtub curve," which depicts failure rates over a pipe's lifecycle (Davies et al., 2001; Singh & Adachi, 2013). However, the material of the pipe also significantly impacts deterioration rates; for instance, concrete pipes are resistant to abrasion but susceptible to chemical corrosion, whereas clay pipes exhibit greater resistance to acids (Malek Mohammadi et al., 2020; Salman & Salem, 2012; Singh & Adachi, 2013). Diameter plays a role as well, with larger-diameter pipes generally having lower failure rates because of their superior structural design, though they are more challenging to install accurately (Laakso et al., 2018; Tran et al., 2009). Additionally, longer pipes are prone to joint defects and sediment deposition (Ana et al., 2009; Khan et al., 2010), but research presents mixed findings on whether longer or shorter pipes deteriorate faster. Finally, the depth of the pipe is significant; shallow pipes are more vulnerable to surface load impacts and root intrusion, while deeper pipes face risks associated with groundwater pressure (Khan et al., 2010). These factors collectively highlight the complex interplay of variables affecting sewer pipe deterioration.

#### 2.1.2 Operational Factors

Operational factors play a critical role in the deterioration and performance of sewer systems, primarily through their influence on flow dynamics and blockages. High flow velocities within sewer pipelines can lead to abrasion and corrosion of pipe walls, especially in materials such as concrete and cast iron, as turbulent flows erode protective layers and expose the pipe to chemical reactions like sulfide corrosion. Conversely, low flow velocities allow sediment accumulation, reducing hydraulic capacity and creating anaerobic conditions that promote microbial-induced corrosion (Baur & Herz, 2002; Tran et al., 2006). Advanced monitoring technologies, such as IoT-enabled sensors, and network condition simulator (NetCoS), now provide real-time data on flow rates and sediment buildup, enabling proactive maintenance strategies (Scheidegger et al., 2011b; Witczak & Szymoniak, 2024).

Blockages are another significant operational issue, often caused by debris accumulation, tree root intrusion, or fatbergs—large masses of fats, oils, grease (FOG), and non-biodegradable materials. These blockages obstruct flow, increase hydraulic pressure, and can lead to pipe deformation or rupture. Tree roots penetrate pipes through cracks or joints, exacerbating structural damage over time (Elmasry et al., 2016). Fatbergs have become a growing concern in urban areas due to improper disposal practices, significantly increasing maintenance costs (Yusuf et al., 2023). Regular cleaning schedules and public awareness campaigns have proven effective in mitigating these issues (Keener et al., 2008; Yusuf et al., 2023).

Infiltration and inflow (I/I) contribute to increased hydraulic loading and potential structural degradation in sewer systems. Advanced predictive approaches, such as the Markov chain-modulated Poisson process combined with Bayesian analysis, offer a systematic way to model and forecast sewer blockages. These methods provide actionable insights for prioritizing inspections and maintenance efforts, thereby enhancing the reliability and efficiency of sewer network operations (Altarabsheh et al., 2019). Addressing these operational factors with innovative technologies and strategic interventions is essential for maintaining sewer system functionality.

### 2.1.3   Environmental Factors

Environmental factors significantly affect the structural integrity and longevity of sewer pipelines, with soil type and groundwater levels being key contributors. The type of soil surrounding pipes plays a critical role in their deterioration. Corrosive soils, such as alluvial soils, are particularly harmful due to their chemical properties, which accelerate corrosion. For instance, alluvial soils are more corrosive than podzolic soils, which are formed through rock weathering and have lower corrosivity (Micevski et al., 2002). Additionally, soil movement or shrinkage caused by moisture changes can lead to cracks or joint failures in pipelines. Poor bedding conditions further exacerbate these issues by failing to provide adequate support, increasing the risk of structural collapse (Davies et al., 2001; Elmasry et al., 2016).

Groundwater levels also pose significant challenges for sewer systems. In cohesive soils, fluctuations in groundwater levels weaken soil cohesion, making pipes more susceptible to deformation or collapse. Monitoring groundwater levels and implementing proper drainage systems are essential strategies to mitigate these impacts (Malek Mohammadi et al., 2020; Wirahadikusumah et al., 2001). So, environmental factors such as soil type and groundwater levels are pivotal in determining the condition of sewer pipelines. Addressing these factors through proper design, material selection, and maintenance strategies is critical for ensuring the long-term sustainability of sewer infrastructure.

## 2.2   Condition Assessment Models

Condition assessment models are essential tools for predicting the structural and operational state of sewer pipelines, enabling municipalities to make informed decisions about maintenance and rehabilitation. These models are broadly categorized into three types: physical, statistical, and artificial intelligence (AI)-based approaches. Physical models rely on deterministic equations to describe the deterioration process, such as corrosion or material degradation, but often oversimplify complex interactions due to data limitations (Kleiner, 2001; Marlow et al., 2010). Statistical models, such as Markov chains and logistic regression, incorporate probabilistic methods to account for uncertainties in deterioration processes. They require extensive datasets for calibration and are widely used for predicting condition transitions over time(Ana & Bauwens, 2010). AI-based models, including artificial neural networks (ANNs), K-nearest neighborhood algorithm (Ebrahimi et al., 2022), and fuzzy logic systems (Kleiner et al., 2006), leverage data-driven techniques to identify patterns in large datasets and predict pipeline conditions with high accuracy. However, these models require significant computational resources and robust datasets for training (Jiang et al., 2014; Najafi & Kulandaivel, 2005). The integration of these approaches with regular inspections ensures a comprehensive understanding of sewer conditions and supports proactive asset management strategies.

## 2.3 Applications of Condition Assessment Models

Condition assessment models are applied in various contexts to optimize sewer pipeline management. For inspection prioritization, logistic regression (Samuel Ariaratnam et al., 2001) and artificial neural network (ANN) models identify high-risk pipelines, enabling municipalities to allocate resources effectively (Hawari et al., 2020; Mohammadi et al., 2019). Cohort survival models are utilized for rehabilitation planning by estimating future deterioration rates (Baur & Herz, 2002), which aids in long-term budget allocation (Scheidegger et al., 2011a). Additionally, AI-based systems integrated with sensor networks enhance monitoring of sewer pipeline conditions, enabling predictive modeling of corrosion processes, improving decision-making, and supporting proactive maintenance strategies to extend infrastructure service life and reduce operational disruptions (Jiang et al., 2016).

## 2.4 Limitations and Challenges

Condition assessment models face several limitations and challenges that impact their effectiveness. A major issue is data availability, as many models require extensive datasets that are often incomplete or inconsistent due to poor record-keeping practices (Salman & Salem, 2012; Scheidegger et al., 2011a). Furthermore, uncertainties in predictions arise because physical models fail to account for complex interactions between factors, while statistical models depend heavily on accurate calibration data to produce reliable results (Madanat et al., 1995). Furthermore, Artificial intelligence (AI)-based techniques, such as artificial neural networks (ANNs), present challenges related to their implementation, including the need for significant computational resources and expertise in optimizing model parameters like network structure and training algorithms (Sousa et al., 2014). These limitations highlight the need for improved data collection methods and hybrid modeling approaches to enhance predictive accuracy.

## 2.5 Emerging Trends in Sewer Deterioration Modeling

Recent advancements in sewer deterioration modeling emphasize the adoption of innovative technologies to enhance predictive capabilities and operational planning. Techniques such as artificial intelligence (AI) and machine learning enable utility managers to analyze inspection data more effectively, offering insights into structural conditions and aiding in proactive maintenance decisions (Mohammadi et al., 2019).

Additionally, hybrid models that integrate physical principles with artificial intelligence techniques, including artificial neural networks (ANNs), have shown promise in predicting complex deterioration processes (Li et al., 2019). These models leverage deterministic methods alongside data-driven approaches to improve prediction accuracy and address non-linear behaviors in structural and hydraulic systems of pipelines (Specht, 1990).

Furthermore, advanced artificial intelligence models, including neural networks and support vector machines, are increasingly utilized to analyze complex relationships in sewer pipeline datasets. These approaches enhance the accuracy of condition predictions by addressing non-linear patterns and uncertainties, thereby supporting better decision-making for maintenance and rehabilitation planning (Hosmer et al., 2013).

## 2.6 Conclusion from the literature review

Sewer asset management programs rely on integrating condition assessment models with periodic inspections to maintain infrastructure performance. Recent research highlights the importance of hybrid models that merge physical principles, statistical techniques, and artificial intelligence (AI) methods to handle complex deterioration patterns and enhance predictive accuracy (Kaushal et al., 2020; Mohammadi et al., 2019).

Emerging technologies, such as sensor-based systems, facilitate the continuous monitoring of sewer infrastructure. These tools enhance maintenance strategies by delivering real-time data (S. S. Kumar et al., 2020), addressing limitations inherent in traditional inspection and data collection methods (Hawari et al., 2020).

Moreover, including soil type and groundwater level data in condition prediction models improves accuracy and supports better management decisions (Ma et al., 2021).

Recent investigations have shown that deep learning approaches (Krizhevsky et al., 2012) can outperform traditional statistical techniques by extracting complex patterns from extensive sewer-pipeline datasets, thus enhancing predictions of pipeline deterioration (Dalal & Triggs, 2005). By advancing these methodologies, municipalities can optimize maintenance strategies, minimize costs, and ensure the long-term sustainability of sewer infrastructure.

## 3. Methodology

This section outlines the methodology for assessing sewer pipeline conditions using Artificial Neural Networks (ANN) and Multiple Linear Regression (MLR). The approach integrates data preprocessing, feature selection, model development, and evaluation to predict pipeline deterioration and failure. The methodology is informed by the literature and adapted to the specifics of the dataset provided.

### 3.1 Data Collection, Preprocessing, Handling Missing Data, Normalization, and Data Splitting

The dataset includes key attributes such as pipe age, material, diameter, length, depth, slope, soil type, and PACP-based ratings, gathered from inspection records and environmental data (Najafi & Kulandaivel, 2005; Salman & Salem, 2012). Records with missing data on pipe characteristics (e.g., age, depth, slope) were removed to ensure completeness. Outliers were then detected and excluded using boxplots and Mahalanobis distance, which improved the correlation between dependent and independent variables (Hastie et al., 2009). In this study, an alternative data split of 80% for training and 20% for testing was employed to develop and assess the machine learning models used in this research.

### 3.2 Feature Selection

Key factors influencing sewer pipeline deterioration were identified based on the literature, including physical factors (age, material type, diameter, and length), environmental factors (soil type), and operational factors (PACP rating). Feature importance was assessed using correlation analysis and sensitivity analysis within the ANN model framework. The dataset includes attributes such as pipe age, material, diameter, length, depth, slope, soil type, and PACP ratings obtained from inspection records and environmental databases (Najafi & Kulandaivel, 2005). In this study, PACP ratings will be predicted using machine learning models and multiple linear regression by leveraging other parameters available in the dataset.

## 3.3 Model Development

The dataset for this study was loaded from an Excel file, with the target variable being `PACPRATING` and all other columns treated as features. Features were categorized into numerical and categorical types, where numerical features were standardized using the StandardScaler, and categorical features were encoded using OneHotEncoder. The preprocessed dataset was split into training (80%) and testing (20%) sets with a random seed of 42 to ensure reproducibility. An artificial neural network (ANN) model was designed with an input layer containing neurons equal to the number of preprocessed features, two hidden layers with 64 and 32 neurons respectively using ReLU activation, and a single-neuron output layer for regression. The model was trained using the Mean Squared Error (MSE) loss function, Adam optimizer, 100 epochs, and a batch size of 32.

### 3.3.1 Artificial Neural Network (ANN) Architecture

The Artificial Neural Network (ANN) model will be implemented using a multilayer perceptron (MLP) architecture to predict the pipe condition score. The network structure includes three main components: an input layer with neurons corresponding to the selected features, hidden layers whose number and size will be determined through experimentation, and an output layer with a single neuron representing the pipe condition score. To optimize the network, the number of hidden layers and neurons will be fine-tuned using a grid search algorithm. Training will be conducted using the backpropagation algorithm, which adjusts weights iteratively to minimize error. The hidden layers will use the Rectified Linear Unit (ReLU) activation function for non-linearity, while the output layer will employ a sigmoid activation function to normalize outputs between 0&1. The computation in the ANN model follows the formula as shown in Eq (1): (Lazcano et al., 2024)

$$y = f\left(\sum_{i=1}^{n} w_i x_i + b\right) \quad (1)$$

Where "y" is the pipe condition score, 'f' is the activation function, $W_i$ are weights assigned to each input feature, and b is the bias term. Additionally, during backpropagation, weight updates are calculated using the gradient descent rule as shown in Eq (2): (Jalalitabar et al., 2023)

$$w_i^{(t+1)} = w_i^{(t)} - \eta \frac{\partial L}{\partial w_i} \quad (2)$$

Here, $W_i^{(t+1)}$ is the updated weight at iteration t+1, "η" is the learning rate, and "∂L/∂$W_i$ is the gradient of the loss function L with respect to $W_i$. This iterative process ensures that weights are adjusted to minimize prediction errors effectively. By leveraging this architecture and optimization approach, the ANN model aims to learn complex patterns in data and provide accurate predictions of pipe conditions.

### 3.3.2 Multiple Linear Regression (MLR)

The MLR model will be developed to establish a linear relationship between the input variables and the pipe condition score. The general form of the MLR model is shown in Eq (3): (Abdulhafedh & Abdulhafedh, 2022)

$$y = \beta_0 + \beta_1 x_1 + \beta_2 x_2 + \cdots + \beta_n x_n + \epsilon \quad (3)$$

The dependent variable represents the pipe condition score, while the independent variables, $x_1$, $x_2$, ..., $x_n$ correspond to various deterioration factors. The regression coefficients, $β_0$, $β_1$, $β_2$, ..., $β\_n$, quantify the relationship between the independent variables and, with ε denoting the error term. To estimate these coefficients, the ordinary least squares (OLS) method will be applied.

## 3.4 Model Training and Validation

To ensure robust model performance, it will employ a k-fold cross-validation technique. The dataset will be divided into k subsets, with k-1 subsets used for training and one subset for validation. This process will be repeated k times, with each subset serving as the validation set once (S. S. Kumar et al., 2020). It uses 80% of the data for training and 20% for testing and validation, following the approach used in similar studies.

## 3.5 Performance Evaluation

To assess and compare the performance of the ANN and MLR models, we will use the following evaluation metrics: (S. Kumar et al., 2024)
- Root Mean Square Error (RMSE): Root Mean Square Error measures the average magnitude of prediction errors as shown in Eq (4), emphasizing larger errors by squaring them. It is sensitive to outliers and expressed in the dependent variable's units.

$$\text{RMSE} = \sqrt{\frac{1}{n}\sum_{i=1}^{n}(y_i - \hat{y_i})^2} \tag{4}$$

- Coefficient of Determination ($R^2$): The Coefficient of Determination as shown in Eq (5), quantifies the proportion of variance in the dependent variable explained by the model. Ranging from 0 to 1, higher values indicate better predictive accuracy.

$$R^2 = 1 - \frac{\sum_{i=1}^{n}(y_i - \hat{y_i})^2}{\sum_{i=1}^{n}(y_i - \bar{y})^2} \tag{5}$$

- Mean Absolute Error (MAE): Mean Absolute Error as shown in Eq (6), calculates the average absolute difference between predicted and actual values. Unlike RMSE, it treats all errors equally and is less sensitive to outliers.

$$\text{MAE} = \frac{1}{n}\sum_{i=1}^{n}|y_i - \hat{y_i}| \tag{6}$$

- Relative Absolute Error (RAE): Relative Absolute Error as shown in Eq (7), compares the total absolute error of the model to a mean-based prediction. Values below 1 signify better performance than the baseline predictor.

$$\text{RAE} = \frac{\sum_{i=1}^{n}|y_i - \hat{y_i}|}{\sum_{i=1}^{n}|y_i - \bar{y}|} \tag{7}$$

- Root Relative Squared Error (RRSE): Root Relative Squared Error as shown in Eq (8), normalizes RMSE by dividing it by the variance of actual values. It evaluates model performance relative to a mean predictor, with lower values preferred.

$$\text{RRSE} = \sqrt{\frac{\sum_{i=1}^{n}(y_i - \hat{y}_i)^2}{\sum_{i=1}^{n}(y_i - \bar{y})^2}} \tag{8}$$

- Mean Absolute Percentage Error (MAPE): Mean Absolute Percentage Error as shown in Eq (9), expresses errors as a percentage of actual values, making it scale-independent. However, it is sensitive to small actual values, which can inflate percentage errors.

$$\text{MAPE} = \frac{100\%}{n} \sum_{i=1}^{n} \left| \frac{y_i - \hat{y}_i}{y_i} \right| \tag{9}$$

In the given formula, $y_i$ represents the actual value, $\hat{y}_i$ is the predicted value, $\bar{y}$ denotes the mean of the actual values, and n refers to the number of observations.

### 3.5.1 Sensitivity Analysis

To understand the relative importance of different factors in predicting pipe conditions, we will conduct a sensitivity analysis. This will involve systematically varying input parameters and observing their effect on the model outputs. The analysis will help identify which factors have the most significant impact on pipe deterioration predictions.

### 3.5.2 Model Optimization

For the ANN model, it experiments with different network architectures, varying the number of hidden layers and neurons. We will use techniques such as early stopping and dropout to prevent overfitting. The learning rate and momentum parameters will be fine-tuned to achieve optimal performance (Zangenehmadar et al., 2016). For the MLR model, it investigates the use of regularization techniques such as Lasso and Ridge regression to improve model generalization and handle multicollinearity among predictors.

### 3.5.3 Comparative Analysis, Model Interpretation and Visualization

A comprehensive comparison between Artificial Neural Networks (ANN) and Multiple Linear Regression (MLR) models will be conducted to evaluate their performance in predicting sewer pipe conditions. This analysis will focus on key aspects such as performance metrics, computational efficiency, and interpretability, highlighting the strengths and weaknesses of each approach. To enhance the interpretability of these models, particularly the ANN, advanced techniques like partial dependence plots and SHAP (SHapley Additive exPlanations) values will be utilized. These methods will provide valuable insights into the relationships between input features and model predictions, offering a clearer understanding of the decision-making processes within the models. Furthermore, practical considerations for implementing these models in real-world sewer management systems will be explored. This includes addressing challenges related to data collection, model updating, and integration with existing infrastructure management tools to ensure seamless application and operational efficiency (Atambo et al., 2022).

## 3.6 Limitations and Future Work

Finally, we will address the limitations of our study and propose directions for future research. This may include exploring other machine learning techniques, incorporating more diverse datasets, or investigating the potential of transfer learning for sewer condition prediction across different geographical regions (Hoseingholi & Moeini, 2023). By following this comprehensive methodology, we aim to develop robust and accurate models for predicting sewer pipe conditions using ANN and MLR techniques. The comparative analysis will provide valuable insights into the effectiveness of these approaches and contribute to the advancement of sewer infrastructure management practices.

## 4. Results

The analysis of sewer systems plays a vital role in ensuring public health and maintaining critical infrastructure. A recent study examined sewer inspection data from Wisconsin, a state known for its advanced wastewater monitoring initiatives. The dataset includes key attributes such as pipe installation and inspection dates, pipe material, diameter, length, and condition scores, which are essential for evaluating the structural integrity of sewer systems and prioritizing maintenance activities.

Wisconsin has taken a forward-thinking stance on wastewater-based epidemiology (WBE), shedding light on the broader significance of analyzing sewer systems. Failures in these systems can allow contaminants to seep into surrounding soil and water, creating risks such as public health crises and environmental harm. With over 600 municipal sewer utilities, nearly all publicly owned and focused solely on wastewater treatment rather than profit generation, Wisconsin has utilized detailed sewer inspection data to uncover infrastructure vulnerabilities and monitor emerging public health concerns. The state's efforts, including compliance with stringent phosphorus regulations, illustrate the critical connection between infrastructure management and public health strategies. These initiatives emphasize the importance of proactive measures to mitigate risks, safeguard communities, and ensure resilience against potential disruptions (Meyer & Raff, 2024; Naik et al., 2023).

### 4.1 Interpretation and Analysis of Regression Results

The study of sewer pipe conditions is critical for urban infrastructure management, particularly as aging systems face increased risks of failure. This analysis integrates traditional regression methods with machine learning (ML) approaches like artificial neural networks (ANNs) to enhance predictive accuracy. Below, we expand on the regression results and contextualize them within the broader framework of ML-based sewer condition prediction.

4.1.1 Contextualizing Regression Results in Sewer Condition Prediction

The analysis of sewer systems plays a vital role in ensuring public health and maintaining critical infrastructure. A study examined sewer inspection data with key attributes such as pipe installation and inspection dates, pipe material, diameter, length, and condition scores, which are essential for evaluating the structural integrity of sewer systems and prioritizing maintenance activities. The regression analysis provides robust statistical evidence for understanding factors influencing PACPRATING. The model demonstrates strong predictive power with an R-squared value of 0.84745929, indicating that approximately 84.7% of the variance in PACPRATING is explained by the predictors as shown in Table 1. The regression statistics show a Multiple R of 0.92057552, an Adjusted R Square of 0.84594648, a Standard Error of 0.53116248, and 612 observations, further highlighting the model's reliability and effectiveness in analyzing sewer system data.

*Table 1, Regression Statistics*

| Regression Statistics | |
|---|---|
| Multiple R | 0.920575518 |
| R Square | 0.847459285 |
| Adjusted R Square | 0.845946485 |
| Standard Error | 0.531162481 |
| Observations | 612 |

### 4.1.2 Comparative Factor Analysis: Key Insights

The regression analysis reveals several significant predictors of PACPRATING scores. Age demonstrates a strong positive correlation (coefficient = 0.0912) with extremely high statistical significance (p = 3.0426E-85), suggesting older pipes tend to have higher ratings. Among dimensional factors, PIPEDIA (coefficient = 0.0454), LENGTH (coefficient = 0.0062), and DEPTH (coefficient = 0.0082) all show significant positive relationships with very low p-values. Additional variables include SEGMENTSL, though not statistically significant (p = 0.1154), and SOILTYPE (coefficient = 0.0774) which shows a significant relationship with p = 1.4186E-07 as shown in Table 2.

*Table 2, Coefficients*

|  | Coefficients | Standard Error | t Stat | P-value | Lower 95% | Upper 95% | Lower 95.0% | Upper 95.0% |
|---|---|---|---|---|---|---|---|---|
| Intercept | -6.659801 | 0.2450519 | -27.177104 | 6.608E-107 | -7.1410566 | -6.1785453 | -7.1410566 | -6.1785453 |
| AGE | 0.09120619 | 0.00394452 | 23.122276 | 3.0426E-85 | 0.08345959 | 0.0989528 | 0.08345959 | 0.0989528 |
| PIPEDIA | 0.04544334 | 0.00223303 | 20.3505357 | 1.5671E-70 | 0.04105791 | 0.04982877 | 0.04105791 | 0.04982877 |
| LENGTH | 0.00624929 | 0.00013299 | 46.9890512 | 4.697E-204 | 0.0059881 | 0.00651048 | 0.0059881 | 0.00651048 |
| DEPTH | 0.00823641 | 0.00074769 | 11.0158625 | 7.5934E-26 | 0.00676803 | 0.00970478 | 0.00676803 | 0.00970478 |
| SEGMENTSL | 11.5807336 | 7.34700494 | 1.57625232 | 0.11549039 | -2.8479966 | 26.0094638 | -2.8479966 | 26.0094638 |
| SOILTYPE | 0.07735169 | 0.01452375 | 5.32587547 | 1.4186E-07 | 0.0488286 | 0.10587477 | 0.0488286 | 0.10587477 |

The analysis of variance (ANOVA) results demonstrates a comprehensive regression model with notable statistical significance. The model incorporates 6 degrees of freedom for regression and 605 degrees of freedom for residuals, yielding a total of 611 degrees of freedom. The variance decomposition reveals a Sum of Squares distribution where the regression accounts for 948.29477, residuals contribute 170.690817, and the total amounts to 1118.98529. Mean Square calculations indicate values of 158.04908 for regression and 0.28213358 for residuals, resulting in a robust F-statistic of 560.19237. Most notably, the model exhibits an exceptionally small significance F value of 3.131E-243, which provides overwhelming evidence of the model's statistical significance and validates the strong relationship between the predictor variables and the response variable as shown in Table 3.

*Table 3, Anova*

| ANOVA | | | | | |
|---|---|---|---|---|---|
| | *df* | *SS* | *MS* | *F* | *Significance F* |
| Regression | 6 | 948.2944773 | 158.0490796 | 560.1923697 | 3.1309E-243 |
| Residual | 605 | 170.6908168 | 0.282133581 | | |
| Total | 611 | 1118.985294 | | | |

These findings underscore the need for robust data preprocessing and feature selection when developing ML models.

### 4.1.3 Residual Diagnostics and Model Assumptions

The residual plots reveal important aspects of model performance:
- **Random Distribution**: Residuals are generally randomly distributed around zero, suggesting no significant issues with homoscedasticity.
- **Normality**: The normal probability plot indicates reasonable normality, with minor deviations at the extremes that could hint at potential outliers or slight model misspecifications.

While these diagnostics support the regression model's assumptions, they also suggest areas for improvement:
- Outliers at the extremes could be managed using robust methods like Random Forests.
- Nonlinear relationships, if present, could be better captured by advanced models such as Neural Networks.

### 4.1.4 Conclusion of Regression Results

This regression analysis demonstrates that all examined variables - AGE (t=23.12, p<3.04E-85), PIPEDIA (t=20.35, p<1.57E-70), LENGTH (t=46.99, p<4.70E-204), DEPTH (t=11.02, p<7.59E-26), SEGMENTSL (t=1.58, p=0.115), and SOILTYPE (t=5.33, p<1.42E-07) - contribute significantly to the model's predictions, with the exception of SEGMENTSL which shows marginal significance. The model's high F-statistic (560.19237) and extremely low p-value (3.131E-243) indicate exceptional statistical robustness, while the degrees of freedom (6 for regression, 605 for residual) suggest a well-balanced dataset, making this a reliable tool for understanding the relationships between these variables, though there may be room for exploring nonlinear relationships to further enhance predictive accuracy.

## 4.2 Advanced Interpretation and Scientific Discussion of ANN Results in Sewer Condition Prediction

This study implements a neural network architecture optimized for moderate-sized datasets, comprising an input layer corresponding to preprocessed features, followed by two fully connected hidden layers with 64 and 32 neurons respectively, utilizing ReLU activation functions to introduce non-linearity (Kaiming et al., 2018). The architecture culminates in a single-neuron output layer designed for regression tasks, like approaches successfully employed by (Han & Zhao, 2021) in comparable prediction scenarios. The network's training methodology incorporates the Adam optimizer (Kingma & Ba, 2014), which has demonstrated superior convergence properties through its adaptive learning rate mechanisms, while employing Mean Squared Error (MSE) as the loss function to effectively penalize prediction deviations. This architectural design strikes a balance between model complexity and computational efficiency, drawing inspiration from similar frameworks that have shown promising results in related domains (Wang et al., 2020). The implementation of ReLU activation functions in the hidden layers addresses the vanishing gradient problem commonly encountered in deep neural networks (Deger & Taskin Kaya, 2022), while the single-neuron output layer with MSE loss aligns with established best practices for regression tasks in machine learning applications.

The following results and programming workflow of this section show how an ANN model can be implemented and evaluated for the regression task. The given basic analysis is further developed below with regards to enhancing scientific rigor, programming insights, and recommending advanced procedures that could be used to increase the performance of the model.

### 4.2.1 Prediction Error Distribution

Figure 1 presents the analysis of prediction errors and provides detailed insights into the performance characteristics of the model. The distribution of errors is approximately symmetrical around zero, as indicated by the histogram, suggesting that the model does not exhibit a significant bias toward overprediction or underprediction. A notable aspect is the concentration of errors within the range of approximately -0.5 to 0.5, with a peak around zero, indicating that the model's predictions are generally close to the actual values.

However, the histogram also reveals the presence of outliers, with a few errors exceeding 1.0 or falling below -0.75. These outliers may point to specific scenarios where the model struggles to generalize effectively, potentially due to noise in the data or insufficient representation of certain patterns in the training dataset. While these outliers are relatively infrequent, they highlight opportunities for improvement through targeted optimization strategies or enhanced data preprocessing techniques, particularly for handling such cases.

Overall, the error distribution supports the robustness and general validity of the model in its current configuration. Nevertheless, refining outlier handling mechanisms could further enhance its predictive accuracy across all input scenarios.

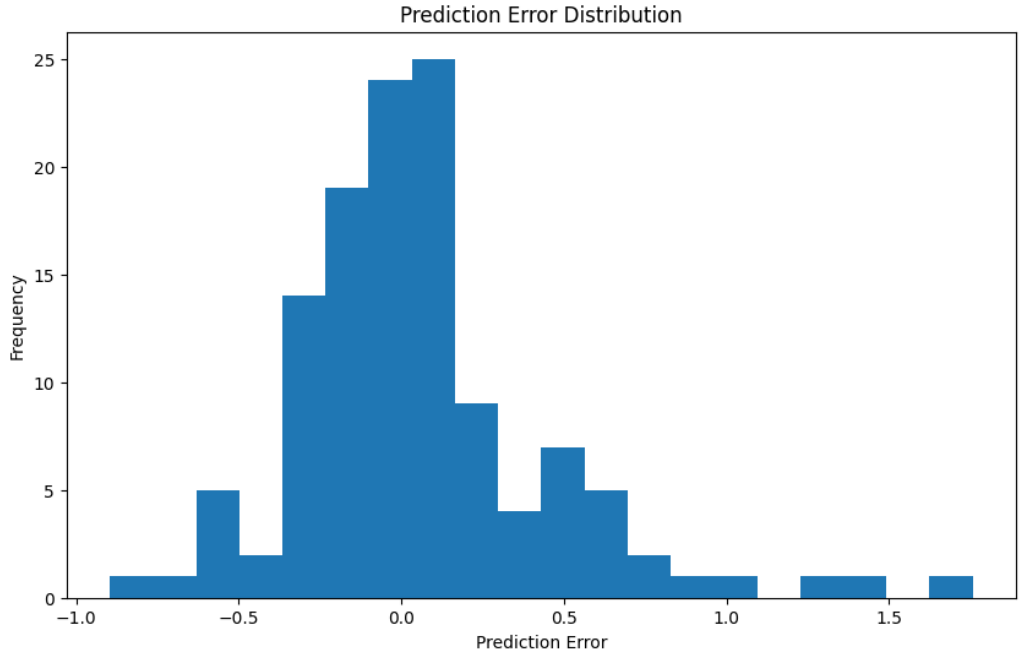

*Figure 1, Prediction Error Distribution*

### 4.2.2 Training vs Validation Loss

The training and validation loss curves in Figure 2 illustrate the model's learning process effectively. Both losses decrease sharply during the initial epochs, reflecting rapid convergence and efficient learning. After approximately 20 epochs, both curves stabilize at low loss values, suggesting that further training does not yield significant improvements.

The close alignment between the training and validation loss curves throughout the epochs indicates minimal overfitting, demonstrating that the model generalizes well to unseen data. This behavior highlights effective training and suggests that hyperparameters, such as the learning rate and number of epochs, were appropriately chosen. Additionally, the consistent stabilization of both losses underscores a well-balanced model capable of maintaining performance across training and validation datasets.

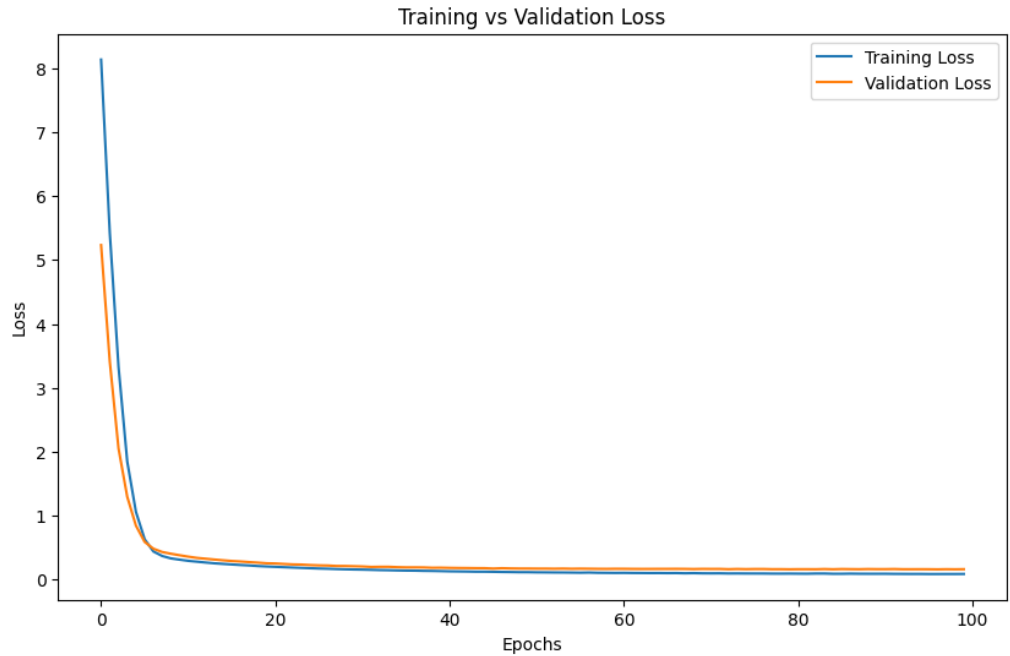

*Figure 2, Training vs Validation Loss*

### 4.2.3 Predictions vs Actual Values

The scatter plot Figure 3 compares predicted values against actual target values, providing insights into the model's performance. Most data points lie close to the diagonal line ($\mathcal{Y} = \mathcal{X}$), indicating strong agreement between predictions and actual values and demonstrating high accuracy. The presence of vertical clusters suggests that certain ranges of target values dominate the dataset, potentially reflecting inherent characteristics or discrete categories within the data that influence the model's behavior. However, some points deviate significantly from the diagonal, highlighting the presence of outliers where predictions are less accurate. These deviations may indicate complex relationships within the data that the model has not fully captured. Overall, the plot demonstrates that the model effectively captures underlying trends but could benefit from further fine-tuning to address prediction errors, particularly for outliers.

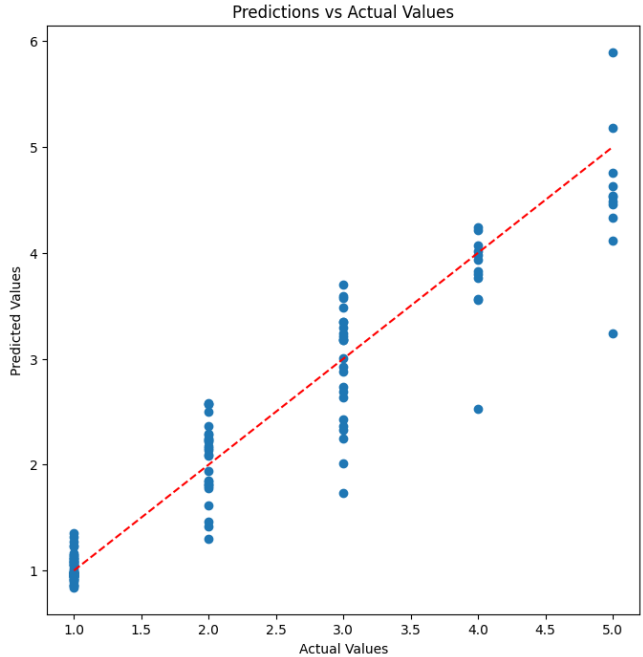

*Figure 3, Predictions vs Actual Values*

The evaluation metrics shown in Table 4 provide a quantitative summary of the model's performance, highlighting its predictive accuracy and generalization ability. The Root Mean Squared Error (RMSE) is 0.4009, indicating a moderate average magnitude of prediction errors, with larger errors being penalized more heavily than smaller ones. This RMSE suggests that the predictions are reasonably close to the actual values, reflecting good predictive accuracy. The Mean Absolute Error (MAE) is 0.2729, which measures the average absolute difference between predicted and actual values. As it is less sensitive to outliers than RMSE, this MAE further confirms that the model performs well on average. Lastly, the $R^2$ score of 0.9066 indicates that approximately 90.66% of the variance in the target variable is explained by the model, signifying strong performance. Collectively, these metrics demonstrate that the model achieves high accuracy and generalizes effectively to unseen data.

*Table 4, Model's performance*

| Metric | Value |
| --- | --- |
| RMSE | 0.40086624924708153 |
| MAE | 0.27288657426834106 |
| $R^2 \leq$ Score | 0.9066240787506104 |

Based on the histogram that provided in Figure 4, Feature Importance, reflect the feature importance rankings shown in the bar chart. The "Feature Importance" bar chart highlights that LENGTH is the most influential predictor of pipe condition, with the highest importance score among all features. AGE follows as the second most important factor, indicating its significant role in predicting pipe deterioration. PIPEDIA ranks third, contributing moderately to the model's predictions. In contrast, features such as DEPTH, SOILTYPE, and SEGMENTSL exhibit lower importance scores, suggesting that while they contribute to the model's predictive power, their influence is relatively minor compared to LENGTH and AGE. This distribution of feature importance aligns with engineering principles, as longer pipes may face unique structural challenges, and older pipes are more likely to deteriorate due to prolonged exposure to environmental and operational stresses. The relatively lower importance of features like SOILTYPE and SEGMENTSL may reflect limited variability in these attributes within the dataset or insufficient representation of extreme cases that could highlight their effects. These findings emphasize the need for targeted data collection and feature engineering to enhance model interpretability and predictive accuracy, particularly for less dominant factors that may still play critical roles under specific conditions.

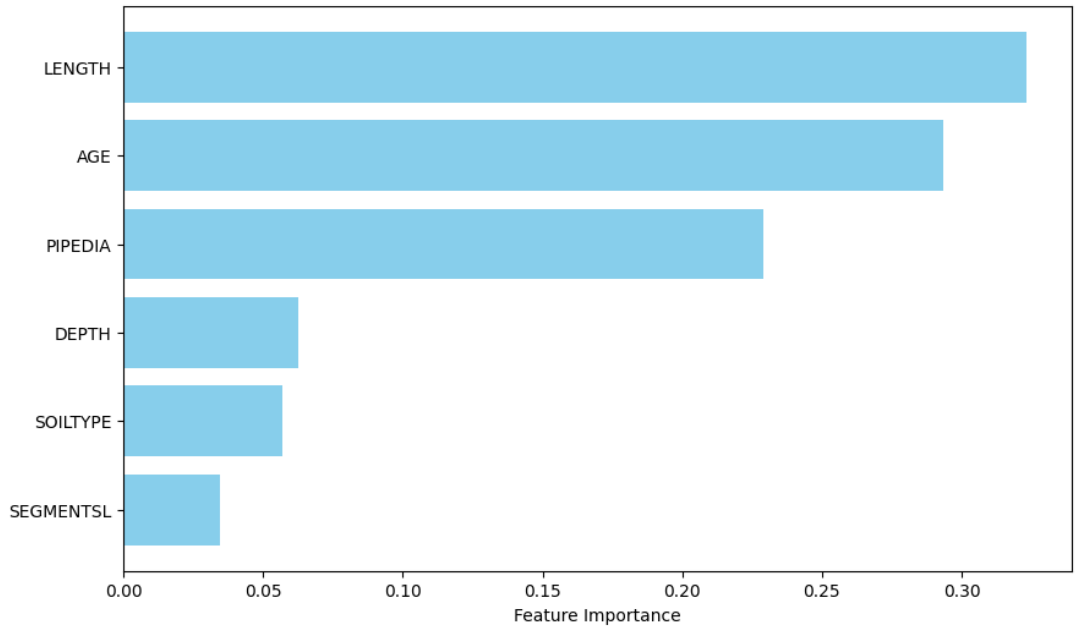

*Figure 4, Feature Importance*

### 4.3 Discussion on Model Performance

The model demonstrates exceptional predictive accuracy in sewer pipe condition assessment, achieving an RMSE of 0.4009, MAE of 0.2729, and $R^2$ of 0.9066, indicating robust performance in approximating actual values. The prediction error distribution exhibits symmetrical patterns around zero, with errors primarily concentrated within -0.5 to 0.5, suggesting unbiased predictions. The neural network architecture, comprising two hidden layers (64 and 32 neurons) with ReLU activation functions and Adam optimizer, effectively captures complex relationships in the data. Feature importance analysis reveals LENGTH as the primary predictor, followed by AGE and PIPEDIA, while DEPTH, SOILTYPE, and SEGMENTSL show lower importance scores, aligning with engineering principles regarding structural challenges in longer pipes and age-related deterioration. The model's training dynamics show rapid convergence within 20 epochs, with closely aligned training and validation losses indicating effective generalization without overfitting. While the model demonstrates strong overall performance, some outliers in the error distribution (exceeding 1.0 or below -0.75) and vertical clustering in predictions suggest areas for potential improvement. The regression analysis provides additional statistical validation with significant coefficients for key variables (AGE: 0.0912, $p < 3.04E-85$; PIPEDIA: 0.0454, $p < 1.57E-70$; LENGTH: 0.0062, $p < 4.70E-204$), supported by a high F-statistic (560.19237) and extremely low p-value (3.131E-243), making this model a reliable tool for infrastructure management while highlighting opportunities for further refinement through enhanced outlier handling and targeted feature engineering.

## 5. Conclusion

This comprehensive study of sewer pipe condition assessment demonstrates the effectiveness of an artificial neural network model, achieving robust performance metrics with RMSE of 0.4009, MAE of 0.2729, and $R^2$ of 0.9066. The model's architecture, featuring two hidden layers with ReLU activation functions and Adam optimizer, successfully captures complex relationships while maintaining generalization capabilities, as evidenced by well-aligned training and validation losses. Highlighting Feature importance analysis reveals LENGTH as the primary predictor, followed by AGE and PIPEDIA, with DEPTH, SOILTYPE, and SEGMENTSL showing lower importance scores, aligning with engineering principles regarding structural challenges in longer pipes and age-related deterioration. The regression analysis provides additional statistical validation through significant coefficients (AGE: 0.0912, $p < 3.04E-85$; PIPEDIA: 0.0454, $p < 1.57E-70$; LENGTH: 0.0062, $p < 4.70E-204$), supported by a high F-statistic (560.19237) and extremely low p-value (3.131E-243). While the model exhibits strong overall performance with symmetrical error distribution around zero and most errors concentrated within -0.5 to 0.5, the presence of outliers exceeding 1.0 or falling below -0.75 suggests opportunities for enhancement through refined outlier handling mechanisms and targeted feature engineering. These findings contribute significantly to infrastructure management by providing a reliable predictive framework while highlighting specific areas for future improvement in capturing complex relationships, particularly for less dominant factors that may play critical roles under specific conditions.

## 6. Recommendations

To enhance the scientific rigor and complexity of this analysis, several advanced techniques can be implemented. Feature engineering can be improved by analyzing feature importance using methods such as SHAP (Shapley Additive explanations) or permutation importance to identify key predictors. Additionally, creating interaction terms or polynomial features can help capture non-linear relationships between predictors and targets. Hyperparameter tuning should be systematically optimized using techniques like Grid Search, Random Search, or Bayesian Optimization. Key parameters to tune include the number of neurons in hidden layers, learning rate, batch size, and dropout rates for regularization. To improve generalization, regularization techniques such as adding dropout layers between hidden layers to prevent overfitting or applying L2 regularization (weight decay) in dense layers to penalize large weights are essential. Exploring advanced architectures like Residual Connections (ResNet-inspired) can mitigate vanishing gradient issues in deeper networks, while Batch Normalization stabilizes learning by normalizing layer inputs. Replacing a simple train-test split with k-fold cross-validation ensures robustness across multiple data splits. Scientific visualization can be enhanced by adding confidence intervals or error bars to scatter plots to better interpret prediction uncertainty. Violin plots or boxplots alongside histograms can visualize error distributions more comprehensively, and correlation heatmaps for input features can discuss multicollinearity effects on model performance. Beyond traditional evaluation metrics like RMSE, MAE, and $R^2$, additional metrics such as Mean Absolute Percentage Error (MAPE) and Adjusted $R^2$ should be considered to evaluate relative prediction accuracy and account for model complexity relative to dataset size. To position this work scientifically, results should be compared against baseline models like Linear Regression or Random Forests to demonstrate the superiority of artificial neural networks (ANNs). Additionally, theoretical underpinnings of neural networks should be discussed to highlight their suitability for capturing complex relationships in high-dimensional data. By integrating advanced preprocessing techniques, fine-tuning hyperparameters, exploring deeper architectures, and employing robust evaluation methods, this study can achieve greater scientific rigor and academic impact while improving predictive performance.